\documentclass{article}

 \usepackage[preprint]{neurips_2025}

\usepackage{wrapfig}
\usepackage{float}
\usepackage{graphicx}
\usepackage{multirow}

\usepackage{microtype}
\usepackage{subfigure}
\usepackage{booktabs} 
\usepackage{bm}
\usepackage{graphicx} 
\usepackage{lipsum} 
\usepackage{amsfonts}
\usepackage{amssymb}
\usepackage{booktabs}
\usepackage{subcaption}
\usepackage{placeins}
\usepackage{tipa}
\usepackage{amsmath}
\usepackage{mathtools}
\usepackage{amsthm}
\usepackage{algorithmic}
\usepackage{algorithm}
\usepackage{hyperref}
\usepackage{changepage}
\usepackage{enumitem}
\usepackage{xcolor}
\usepackage{xspace}

\usepackage{wrapfig}
\usepackage{graphicx}

\usepackage[capitalize,noabbrev]{cleveref} 
\crefname{algorithm}{Algorithm}{Algorithms}
\usepackage[textsize=tiny]{todonotes}

\theoremstyle{plain}
\newtheorem{theorem}{Theorem}

\theoremstyle{definition}

\newtheorem{assumption}{Assumption}
\theoremstyle{remark}

\usepackage[utf8]{inputenc} 
\usepackage[T1]{fontenc}    
\usepackage{hyperref}       
\usepackage{url}            
\usepackage{booktabs}       
\usepackage{amsfonts}       
\usepackage{nicefrac}       
\usepackage{microtype}      
\usepackage{xcolor}         
\usepackage{amsmath, amsthm, amssymb}

\title{Kuramoto-FedAvg: Using Synchronization Dynamics to Improve Federated Learning Optimization under Statistical Heterogeneity}

\author{%
    Aggrey Muhebwa \thanks{Equal Contribution.} \\
    Stanford University\\
    \texttt{amuhebwa@stanford.edu} \\
    \And
    Khotso Selialia \footnotemark[1] \\
    University of Massachusetts Amherst \\
    \texttt{kselialia@umass.edu} \\
    \AND
    Fatima Anwar \\
    \texttt{ fanwar@umass.edu} \\
    University of Massachusetts Amherst \\
    \And
    Khalid K. Osman \\
    Stanford University \\
    \texttt{osmank@stanford.edu} \\
}

\begin{document}

\maketitle

\begin{abstract}
Federated learning on heterogeneous (non-IID) client data experiences slow convergence due to client drift. To address this challenge, we propose Kuramoto-FedAvg, a federated optimization algorithm that reframes the weight aggregation step as a synchronization problem inspired by the Kuramoto model of coupled oscillators. The server dynamically weighs each client's update based on its phase alignment with the global update, amplifying contributions that align with the global gradient direction while minimizing the impact of updates that are out of phase. We theoretically prove that this synchronization mechanism reduces client drift, providing a tighter convergence bound compared to the standard FedAvg under heterogeneous data distributions. Empirical validation supports our theoretical findings, showing that Kuramoto-FedAvg significantly accelerates convergence and improves accuracy across multiple benchmark datasets. Our work highlights the potential of coordination and synchronization-based strategies for managing gradient diversity and accelerating federated optimization in realistic non-IID settings.
\end{abstract}

\section{Introduction}
\label{sec:introduction}
In Federated Learning (FL), a global machine learning (ML) model is collaboratively trained across multiple decentralized clients by aggregating their local updates without requiring the exchange of individual client data\cite{mcmahan2017communication, yang2019federated}. While FL enables privacy-preserving collaboration across various devices (e.g., mobile phones, edge sensors), it introduces new challenges, especially when client data is non-identically distributed (non-i.i.d). Specifically, statistical heterogeneity can lead to client drift, where each client’s local gradient diverges from the global gradient, resulting in unstable and slow convergence\cite{seo2025understanding, karimireddy2020scaffold, zhao2018federated}.

To address client drift introduced by data heterogeneity, several federated optimization techniques have been proposed, each aiming to stabilize training and improve convergence. FedProx\cite{li2020federated} incorporates proximal regularization into each client’s loss to penalize deviations from the global model, which limits excessive drift and results in more stable convergence. However, this introduces a constraint that depends on hyperparameters which must be carefully tuned to balance convergence speed and stability. SCAFFOLD\cite{karimireddy2020scaffold} employs control variates (stochastic correction terms) to offset drift in local updates and stabilize federated optimization. Although this achieves faster convergence than FedAvg\cite{mcmahan2017communication} in non-IID settings, it comes with increased client-side memory requirements and communication overhead. Similarly, techniques such as FedNova\cite{wang2020tackling} correct for objective inconsistencies caused by varying client update frequencies, while FedBN\cite{li2021fedbn} addresses specific heterogeneity in the feature space by keeping local batch normalization parameters separate. Although these techniques improve robustness to non-IID data, they mainly focus on local corrections and modifications, treating each client’s update independently rather than directly addressing the root cause of client drift: the misalignment of gradient directions across clients.
In this work, we propose a novel algorithm to address client drift by framing FL aggregation as a synchronization problem. Our algorithm, Kuramoto-FedAvg, draws inspiration from the classical Kuramoto model, a mathematical framework for synchronizing coupled oscillators\cite{kuramoto1975self, acebron2005kuramoto}. In the traditional Kuramoto formulation, each oscillator is defined by its natural frequency and phase, and synchronization emerges spontaneously when the coupling strength exceeds a critical threshold, aligning the phases of oscillators despite their inherent frequency differences. We consider each client’s model update as an oscillator defined by its direction (phase), with global aggregation acting as a synchronization process. Specifically, at each communication round, the central server measures the phase alignment by calculating the cosine similarity between each client’s local update and the aggregated global gradient, dynamically increasing the contribution of updates that are well-aligned (in-phase) while decreasing the contribution of those that are poorly aligned (out-of-phase). This synchronization-based aggregation coordinates client updates to minimize misalignment, directly addressing gradient diversity at the aggregation stage instead of imposing constraints (such as control variates or penalty terms) during local optimization.

We provide both theoretical and empirical evidence that Kuramoto-FedAvg improves federated optimization under various realistic heterogeneity conditions. Theoretically, we prove that the phase-alignment mechanism reduces client drift and yields a tighter convergence bound than classical FedAvg in heterogeneous settings. Empirically, we demonstrate the advantages of our technique on standard FL benchmarks. Using Convolutional Neural Network (CNN) models for image classification, we show that Kuramoto-FedAvg converges faster and achieves higher accuracy than FedAvg or FedProx. These results support our hypothesis that synchronizing client updates can effectively mitigate gradient diversity and accelerate convergence in FL under non-IID conditions.

In summary, our main contributions are:
\begin{itemize}
    \item \textbf{Synchronization-Based Aggregation}: We propose Kuramoto-FedAvg, a novel federated learning algorithm that addresses gradient misalignment in heterogeneous settings.
    \item \textbf{Theoretical Convergence Analysis}: We provide a theoretical analysis of Kuramoto-FedAvg, demonstrating that the synchronization-based aggregation mechanism reduces client drift and achieves a tighter convergence bound compared to standard FedAvg under non-i.i.d. data distributions.
    \item \textbf{Empirical Validation}: We evaluate Kuramoto-FedAvg using standard federated learning benchmarks and demonstrate significantly faster convergence and improved accuracy compared to baseline methods thereby confirming the practical benefits of synchronization-based coordination in federated optimization
\end{itemize}

\section{Background and Related Work}
\label{sec:background}
\textbf{
Synchronization and The Kuramoto Model}: Synchronization is a fundamental emergent phenomenon in biological, physical, and engineered systems, such as the coordinated flashing of fireflies, the stable frequency alignment in power grids, and the collective behavior of coupled Josephson junctions\cite{strogatz2004sync, wiesenfeld1996synchronization, ermentrout1991adaptive}. The Kuramoto model has become a widely accepted framework to analyze such collective dynamics, offering a mathematically tractable yet powerful description of how global phase coherence arises from local interactions among heterogeneous oscillators \cite{kuramoto1975self, acebron2005kuramoto, rodrigues2016kuramoto}. The Kuramoto model describes the synchronization of $N$ coupled oscillators, each defined by a phase $\theta$ that evolves according to its natural frequency $\omega_i$ and interactions with other oscillators.
\[
    \dot{\theta}_i = \omega_i + \frac{K}{N} \sum_{j=1}^{N} \sin(\theta_j - \theta_i),
\]
where $K$ is the coupling strength and $\theta_i$ are the phases of the oscillators. The Kuramoto model proves that with sufficient coupling strength, oscillators can spontaneously synchronize their phases despite different natural frequency, illustrating a classical second-order phase transition from incoherence to synchronization\cite{crawford1994amplitude, strogatz2000kuramoto}. Recent research has extended the Kuramoto framework beyond the original all-to-all coupling assumption to accommodate complex network topologies, where oscillators interact through heterogeneous, sparse, and often weighted or directed graphs. These generalized Kuramoto models have shown that network topology plays an important role in determining the onset, stability, and qualitative characteristics of synchronization. Structural features such as degree heterogeneity, spectral properties (e.g., spectral gap), community structure, and clustering can significantly influence the critical coupling strength needed for synchronization, the proportion of oscillators that achieve phase coherence, and the resilience of the synchronized state to perturbations\cite{arenas2008synchronization, rodrigues2016kuramoto, skardal2020higher}. Furthermore, analytical tools, including mean-field approximations, spectral analysis, and stability criteria such as the Master Stability Function (MSF) for graph Laplacians, have been developed to investigate these effects in both finite and infinite-sized networks. These advancements have enabled applications to real-world systems, including power grids, where synchronization governs frequency stability and resilience \cite{dorfler2012synchronization, dorfler2013novel}, and neuronal systems, where phase coordination underlies cognition and signal propagation\cite{coutinho2013kuramoto, lopes2016synchronization, pietras2019network}.

\textbf{Client Drift in Federated Learning}:In FL, a centralized ML model is trained across decentralized clients without the need to collect client data centrally. Federated optimization methods, such as FedAvg\cite{mcmahan2017communication}, do not have a finite-time convergence guarantee under general heterogeneity and experience convergence degradation due to non-IID client data, commonly referred to as client drift\cite{seo2025understanding, karimireddy2020scaffold, zhao2018federated}. Client drift refers to the deviation of the local client gradient from the global (centralized) gradient direction, leading to persistent bias in the aggregated update and resulting in slow convergence. Existing methods address client drift by adjusting optimization or aggregation strategies. FedProx\cite{li2020federated} introduces a proximal term in local client objectives, which keeps client updates close to the global model, thereby reducing client drift. Similarly, SCAFFOLD\cite{karimireddy2020scaffold} incorporates variance reduction through control variates to correct client drift, resulting in faster convergence and improved stability. FedNova\cite{li2021fedbn} addresses objective inconsistencies arising from variations in client update steps by employing a normalized averaging technique, ensuring global aggregation aligns with a consistent objective. FedBN\cite{li2021fedbn} adapts local batch normalization at each client to accommodate heterogeneity in data distributions, thereby minimizing feature-space drift. While these techniques mitigate client drift through local constraints or static normalization schemes, they lack mechanisms for dynamically coordinating the global alignment of client updates.


\section{Kuramoto-FedAvg Framework} 
\label{sec:methodology}

In Federated Learning (FL), a central challenge is \emph{client drift}, where stochastic gradients computed on heterogeneous (non-IID) data diverge directionally, resulting in suboptimal and unstable global updates. We address this issue by introducing a novel aggregation mechanism inspired by synchronization dynamics, specifically adapting ideas from the Kuramoto model of coupled oscillators.

\subsection{Phase-Based Aggregation Dynamics}

At the beginning of a given communication round \( t \), let \( w^t \in \mathbb{R}^d \) denote the global model maintained by the server. Each client \( k \in \{1, \dots, K\} \) receives this model and computes an updated local model \( w_k^t \) by performing several steps of local optimization on its private dataset. The local update direction is defined as \( \Delta w_k^t = w_k^t - w^t \), representing the deviation from the current global model.

In standard Federated Averaging (FedAvg), the server aggregates updates using a weighted sum:
\begin{equation}
w^{t+1} = w^t + \sum_{k=1}^K p_k \Delta w_k^t,
\label{eq:fedavg}
\end{equation}
where \( p_k \geq 0 \) and \( \sum_k p_k = 1 \) are aggregation weights, typically proportional to the size of each client's local dataset.

However, FedAvg assumes a level of alignment among update directions that does not hold under non-IID conditions. To address this misalignment, we introduce a phase-based reweighting mechanism that amplifies updates aligned with the dominant direction and suppresses conflicting ones.

\begin{scriptsize}
\begin{algorithm}[t]
\caption{\textsc{Kuramoto-FedAvg}: Federated Learning with Phase-Based Synchronization}
\label{alg:kuramoto-fedavg}
\begin{algorithmic}[1]
\STATE \textbf{Input:} Clients $K$, initial model $w^0$, learning rate $\eta$, total rounds $T$, local epochs $E$, batch size $B$, sync strength $\kappa_0$, decay rate $\beta$
\STATE \textbf{Output:} Final global model $w^T$
\FOR{$t = 0$ to $T-1$}
    \STATE Server broadcasts global model $w^t$ to all clients
    \FOR{each client $k \in \{1,\dots,K\}$ \textbf{in parallel}}
        \STATE Initialize $w_k^t \leftarrow w^t$
        \FOR{$e = 1$ to $E$}
            \STATE Sample minibatch $\mathcal{B}_k \subset D_k$ of size $B$
            \STATE $g_k^e = \nabla \mathcal{L}_k(w_k^{e-1}; \mathcal{B}_k)$
            \STATE $w_k^e = w_k^{e-1} - \eta \cdot g_k^e$
        \ENDFOR
        \STATE $\Delta w_k^t = w_k^E - w^t$
        \STATE Send $\Delta w_k^t$ to server
    \ENDFOR
    \STATE Compute mean update: $\bar{\Delta}^t = \sum_{k=1}^K p_k \cdot \Delta w_k^t$
    \FOR{each client $k \in \{1,\dots,K\}$}
        \STATE Compute phase: $\theta_k^t = \arccos\left( \frac{ \langle \Delta w_k^t, \bar{\Delta}^t \rangle }{ ||\Delta w_k^t|| \cdot ||\bar{\Delta}^t|| } \right)$
    \ENDFOR
    \STATE Compute mean phase: $\bar{\theta}^t = \frac{1}{K} \sum_{k=1}^K \theta_k^t$
    \FOR{each client $k \in \{1,\dots,K\}$}
        \STATE Compute sync weight: $\rho_k^t = \frac{ \sin(\bar{\theta}^t - \theta_k^t) }{ \sum_{j=1}^K \sin(\bar{\theta}^t - \theta_j^t) }$
    \ENDFOR
    \STATE Aggregate update: $w^{t+1} = w^t + \kappa_t \cdot \sum_{k=1}^K \rho_k^t \cdot \Delta w_k^t$
\ENDFOR
\end{algorithmic}
\end{algorithm}
\end{scriptsize}

\subsection{Kuramoto-Inspired Synchronization for Aggregation}

Drawing from the Kuramoto model, which describes the spontaneous synchronization of coupled oscillators via their phase differences, we reinterpret each client update \( \Delta w_k^t \) as an oscillator characterized by its phase angle. This phase, denoted \( \theta_k^t \), captures the direction of the client's update in the parameter space. Let us define the phase as:
\begin{equation}
\theta_k^t = \arccos\left( \frac{ \langle \Delta w_k^t, \bar{\Delta}^t \rangle }{ ||\Delta w_k^t|| \cdot ||\bar{\Delta}^t|| } \right),
\end{equation}
where \( \bar{\Delta}^t = \sum_{k=1}^K p_k \Delta w_k^t \) is the weighted average update direction, and \( \langle \cdot, \cdot \rangle \) denotes the inner product. This angle quantifies how aligned each client's update is with the aggregate direction.

We then define a synchronization-driven weight for each client update based on the phase difference:
\begin{equation}
\rho_k^t = \frac{ \sin(\bar{\theta}^t - \theta_k^t) }{ \sum_{j=1}^K \sin(\bar{\theta}^t - \theta_j^t) },
\end{equation}
where \( \bar{\theta}^t \) is the mean phase of all clients. Intuitively, clients whose updates are more in-phase (i.e., more closely aligned with the global direction) receive higher weight, while those whose updates deviate receive proportionally less influence.

\subsection{Synchronized Aggregation Rule}

Incorporating these dynamic weights, the server updates the global model using:
\begin{equation}
w^{t+1} = w^t + \sum_{k=1}^K \rho_k^t \Delta w_k^t.
\label{eq:kuramoto_fedavg}
\end{equation}

This formulation retains the simplicity of weighted averaging while introducing a coordination mechanism that modulates each client's influence according to its directional coherence. The resulting update dynamically adapts to the geometry of gradient directions across clients, promoting synchronization without requiring explicit client-to-client communication or additional gradient correction terms.

\subsection{Interpretation}

The Kuramoto-FedAvg update in Eq.~\eqref{eq:kuramoto_fedavg} represents a principled deviation from traditional aggregation schemes. Rather than averaging all client updates indiscriminately, it prioritizes updates that reinforce the dominant optimization trajectory. This alignment-driven aggregation reduces the detrimental impact of client drift, especially under high data heterogeneity. As synchronization emerges across training rounds, the global model converges more steadily, with empirically observed improvements in both speed and stability.

\section{Theoratical Analysis}
\label{theoretical_analysis}
\setcounter{assumption}{0}
\begin{assumption}[Federated Objective]
Let the global objective be defined as:
\[
F(w) = \sum_{k=1}^{N} p_k F_k(w),
\]
where each local function \( F_k(w) \) is associated with client \(k\) and \(p_k \geq 0\) such that \( \sum_{k=1}^{N} p_k = 1 \).
\end{assumption}

\begin{assumption}[Smoothness and Unbiased Gradients]
Each local function \( F_k \) is \(L\)-smooth, i.e.,
\[
F_k(v) \leq F_k(w) + \nabla F_k(w)^\top (v - w) + \frac{L}{2} ||v - w||^2.
\]
Also, the stochastic gradient \( g_k^t \) computed on client \(k\) satisfies:
\[
\mathbb{E}[g_k^t] = \nabla F_k(w_t), \quad \text{and} \quad \mathbb{E}||g_k^t - \nabla F_k(w_t)||^2 \leq \sigma_k^2.
\]
Let \( \sigma^2 := \sum_{k=1}^N p_k^2 \sigma_k^2 \).
\end{assumption}

\begin{assumption}[Gradient Diversity]
We define the gradient diversity (client drift) at round \(t\) as:
\[
\Gamma(t) := \sum_{k=1}^N p_k ||\nabla F_k(w_t)||^2 - ||\nabla F(w_t)||^2.
\]
\end{assumption}

\begin{theorem}[FedAvg Convergence Bound]
\label{theorem1}
Let \( w_{t+1} = w_t - \eta_t \sum_{k=1}^{N} p_k g_k^t \) be the global model update in one round of Federated Averaging.
Under Assumptions 1--3, we have:
\[
\mathbb{E}[F(w_{t+1})] \leq F(w_t) - \eta_t ||\nabla F(w_t)||^2 + \frac{L \eta_t^2}{2} \Gamma(t) + \frac{L \eta_t^2}{2} \sigma^2.
\]
\end{theorem}

A complete proof is deferred to Appendix~\ref{app:proof-theorem1}

The standard FedAvg algorithm is susceptible to client drift, where heterogeneous (non-IID) data distributions lead to misaligned local updates, resulting in conflicting contributions that degrade aggregation quality and slow global convergence. Kuramoto-FedAvg introduces a synchronization mechanism inspired by the Kuramoto model, where phase angles capture the alignment between client updates and the global model. Clients with similar update directions (i.e., phase alignment) are given greater influence during aggregation, while misaligned clients contribute less or are dynamically adjusted through coupling forces. This approach promotes faster and more stable global convergence.

\textbf{Theorem 2}: Kuramoto-FedAvg converges faster than standard Fed-Avg under data heterogeneity.
Consider a federated learning objective $F(w) = \sum_{k=1}^{N} p_k F_k(w)$
where each local function $F_k(w)$ is $L-$smooth and $\mu$ strongly convex. If we assume that the dataset is heterogenous and non-IID then we can define the gradient drift at round $t$ as:
\[
\Gamma(t) := \sum_{k=1}^N p_k ||\nabla F_k(w_t)||^2 - ||\nabla F(w_t)||^2.
\]

In standard FedAvg, the convergence bound is given by:
\[
\mathbb{E}[F(w_{t+1})] \leq F(w_t) - \eta_t ||\nabla F(w_t)||^2 + \frac{L \eta_t^2}{2} \Gamma(t) + \frac{L \eta_t^2}{2} \sigma^2.
\]

This shows that the higher the $\Gamma(t)$ i.e., greater client drift and subsequently, the lower the convergence. In FL, this convergence is defined as the number of iterations it takes before the global model converges.

In Kuramoto-FedAvg, each client is assigned a phase $\theta_{k}(t)$, which corresponds to the model's local gradient direction. The synchronization-based weight is thus defined as;
\[
\rho_k(t)=\frac{\sin\bigl(\bar{\theta}(t)-\theta_k(t)\bigr)}{\sum_{j=1}^{N}\sin\bigl(\bar{\theta}(t)-\theta_j(t)\bigr)},
\]
where \(\bar{\theta}(t)\) is the mean phase across clients (the phase of the global client at round $t$).

The client drift is thus represented as:
\[
\Gamma(t) := \sum_{k=1}^N \rho_k(t)^2 \ ||\nabla F_k(w_t)||^2 - ||\nabla F(w_t)||^2.
\]

Based on the Kuramoto model, we know that clients with large phase differences(highly mis-aligned) receive lower weights. Then, we have:
\( \Gamma_{\mathrm{Kuramoto}}(t)<\Gamma(t). \)
Thus, the convergence bound for Kuramoto-FedAvg becomes
\begin{equation}
\mathbb{E}[F(w_{t+1})] \leq F(w_t) - \eta_t ||\nabla F(w_t)||^2 + \frac{L \eta_t^2}{2} \Gamma_{\mathrm{Kuramoto}}(t) + \frac{L \eta_t^2}{2} \sigma^2.
\end{equation}

Therefore, we can assume that for any target error $\epsilon>0$, the number of iterations (or communication rounds with the central server) required by Kuramoto-FedAvg, \(
T_{\mathrm{Kuramoto\mbox{-}FedAvg}}(\epsilon,\Gamma),
\)
is strictly smaller than that of standard FedAvg.
\[
T_{\mathrm{Kuramoto\mbox{-}FedAvg}}(\epsilon,\Gamma) < T_{\mathrm{FedAvg}}(\epsilon,\Gamma).
\]

See Appendix~\ref{app:proof-theorem2} for the proof.

\section{Evaluation}
\label{sec:evaluation}
In this section, we evaluate the effectiveness of our proposed federated optimization technique in synchronizing local gradients across clients with heterogeneous data and improving average model performance (e.g., test accuracy).

\subsection{Experimental Setup}
\label{subsec:experimental_setup}
We conduct experiments on image classification tasks to evaluate the effectiveness of Kuramoto-FedAvg in synchronizing gradients across clients with heterogeneous data while improving overall model utility. Our experiments use three standard benchmark datasets: MNIST \cite{cohen2017emnist}, Fashion-MNIST(FMNIST) \cite{xiao2017fashion}, and CIFAR-10 \cite{recht2018cifar}. For CIFAR-10, we employ a VGGNet-based convolutional model~\cite{tammina2019transfer}, while for MNIST and Fashion-MNIST, we use a standard four-layer CNN~\cite{o2015introduction}. Each dataset is partitioned across $K \in \{10, 20\}$ clients using a label-based non-IID partitioning strategy. Specifically, we implement label sharding by sorting the dataset by class label, dividing it into $K \cdot s$ equal-sized shards, and randomly assigning $s$ shards to each client without replacement. We vary the number of shards per client with $s \in \{3, 5, 10\}$ to control the degree of statistical heterogeneity, ensuring all clients receive the same amount of data but with different label distributions.

Training is conducted over $100$ global communication rounds. In each round, every client performs $2$ local epochs of training with a batch size of $64$. We use stochastic gradient descent (SGD) with a momentum of 0.9, an initial learning rate of 0.01, and a cosine learning rate decay schedule. To enforce gradient synchronization across clients, Kuramoto-FL applies a discretized Kuramoto model to the local gradients, introducing a coupling mechanism characterized by a strength parameter $\kappa \in \{0.1, 0.5, 1.0\}$. This dynamic allows each client's gradient direction to gradually align with that of its neighbors, encouraging coherence before aggregation.

We evaluate Kuramoto-FedAvg using two main metrics. First, we report average test accuracy across all clients to assess overall model utility. Second, we measure synchronization quality by computing the variance of train loss across clients, $\text{Var}_h(\{a^k\}_{k=1}^K)$, where $a^k$ denotes the model's loss on client $k$'s training data.

Furthermore, We compare Kuramoto-FedAvg with two baseline approaches. The first is FedAvg~\cite{li2020federated}, which represents the standard federated optimization strategy. The second is SCAFFOLD~\cite{karimireddy2020scaffold}, which aims to mitigate client-drift by incorporating variance-reducing control variates. Both baselines are implemented using the same data partitions, models, and training protocols to ensure fair comparisons.

Although our experimental scope is limited to a small number of datasets and client counts, we note that the theoretical guarantees established in Section~\ref{theoretical_analysis} ensure the generalizability of Kuramoto-FedAvg's synchronization to broader FL settings.

\subsection{Main Results}
\begin{figure}[h]
  \centering
  \includegraphics[width=1.0\textwidth]{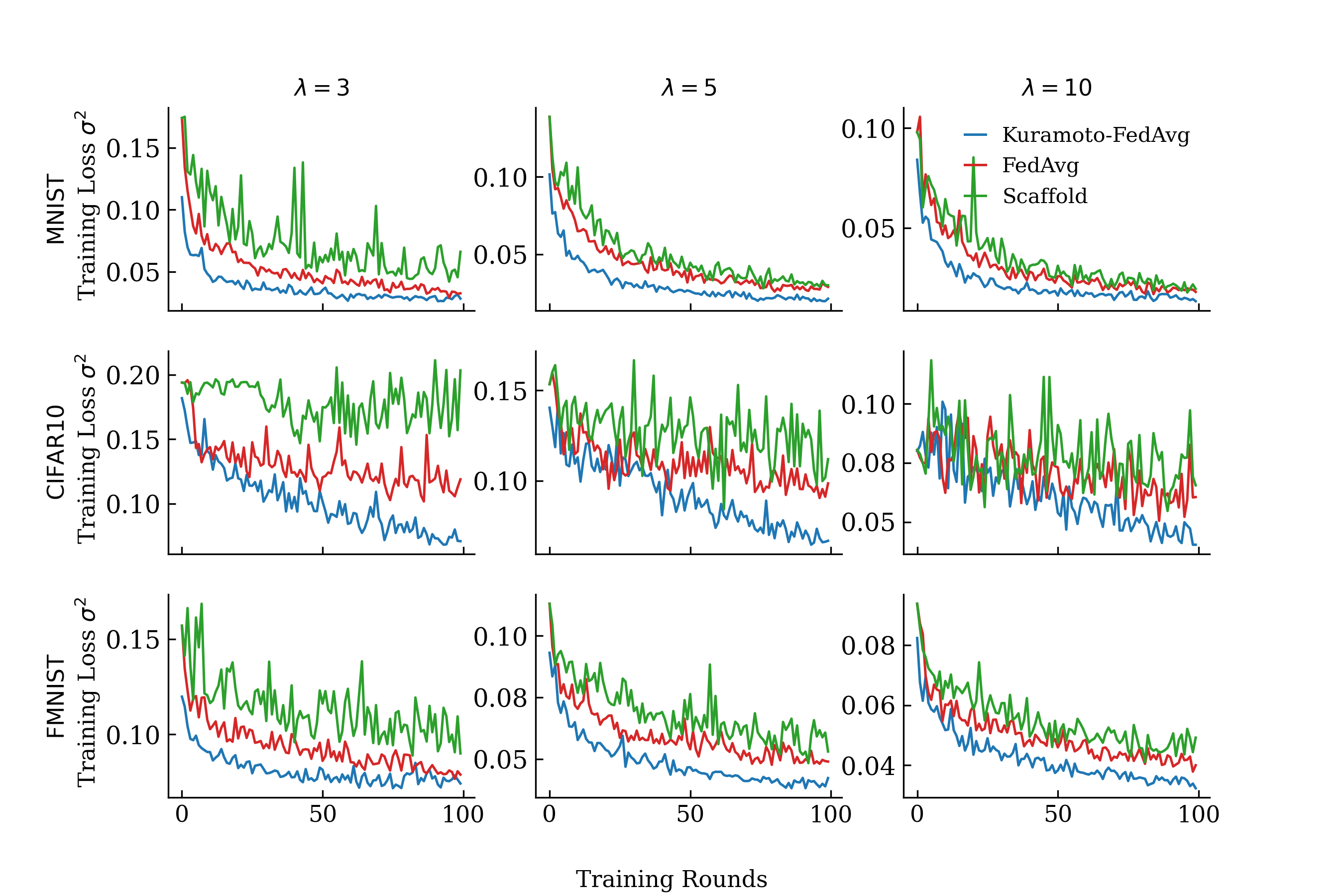}
  \caption{Comparison of Kuramoto-FedAvg with standard FedAvg and SCAFFOLD under varying levels of statistical heterogeneity (controlled by the number of data shards per client, $s \in \{3, 5, 10\}$. A lower variance indicates better synchronization among local client models. Across all benchmark datasets, Kuramoto-FedAvg converges faster and consistently maintains lower variance throughout training.}
  \label{fig:kuramoto_variance}
\end{figure}

We examine the effectiveness of our proposed algorithm in reducing client drift by analyzing the variance of the global model’s losses across training iterations Figure \ref{fig:kuramoto_variance}. A lower variance indicates greater synchronization of model updates among clients, which is synonymous with stable convergence behavior. 

\vspace{0.1em}

Across MNIST and FMNIST, a standard CNN model optimized with Kuramoto-FedAvg rapidly achieves client synchronization compared to FedAvg and SCAFFOLD, particularly under extreme data heterogeneity ($ \lambda = 3$). Kuramoto-FedAvg reaches stable variance levels approximately 20 to 40 rounds earlier than both SCAFFOLD and FedAvg, demonstrating substantial improvements in convergence speed and stability. Under moderate heterogeneity ($ \lambda = 5$) and near-IID conditions ($\lambda = 10$), Kuramoto-FedAvg consistently shows rapid initial synchronization, although the performance gaps narrow slightly due to less severe drift. The benefits of global synchronization are more pronounced on CIFAR-10 when employing a deeper VGGNet-based convolutional architecture. FedAvg and SCAFFOLD struggle with high variance and unstable convergence under severe heterogeneity whereas Kuramoto-FedAvg achieves sustained variance reduction, halving the variance of FedAvg within the first 25 rounds and maintaining consistently low levels thereafter. Under less heterogeneous conditions, Kuramoto-FedAvg continues to outperform baseline methods, albeit with a slightly smaller margin due to decreased client drift. These benefits are further reinforced in Figure \ref{fig:kuramoto_accuracy}, which illustrates the convergence advantages of Kuramoto-FedAvg on the test accuracies at the end of each training round. Kuramoto-FedAvg’s early convergence to higher accuracy levels demonstrates significant efficiency benefits, which are particularly important in resource-constrained federated scenarios.

\begin{figure}[h]
 \centering
  \includegraphics[width=0.80\textwidth]{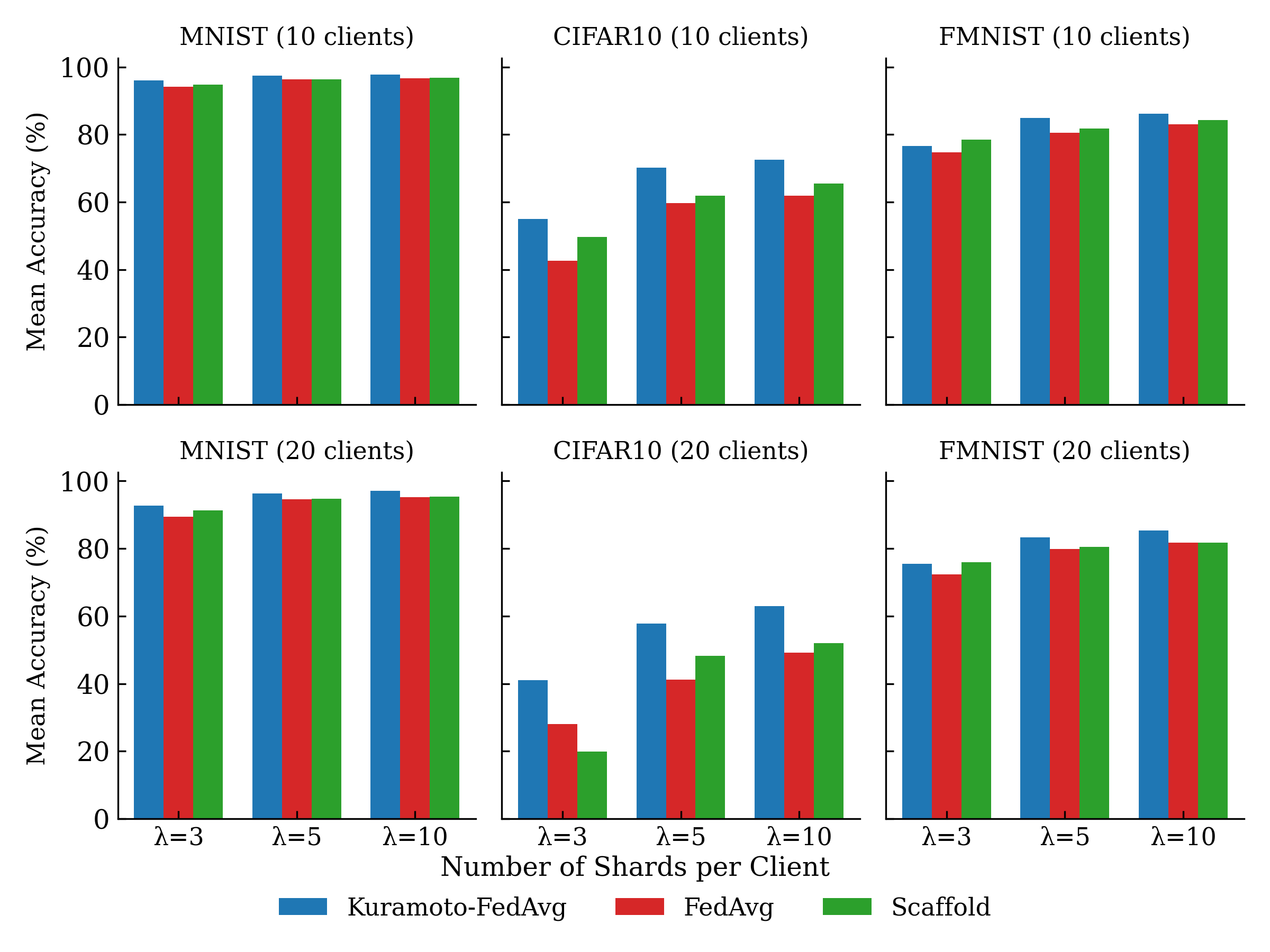}
  \caption{Mean test accuracy across varying degrees of statistical heterogeneity. While all methods achieve higher accuracy with reduced heterogeneity, Kuramoto-FedAvg consistently outperforms the baselines across all heterogeneity settings and datasets.}
  \label{fig:kuramoto_mean_accuracy}
\end{figure}
\vspace{0.1em}

Figure \ref{fig:kuramoto_mean_accuracy} summarizes the overall model performance in terms of mean final accuracy across varying levels of heterogeneity. Kuramoto-FedAvg consistently achieves the highest accuracy across all datasets and heterogeneity levels. Under highly non-IID conditions, the accuracy advantage is more pronounced, with Kuramoto-FedAvg outperforming FedAvg by a notable 10–15\% improvement in accuracy on CIFAR-10, and consistently outperforms SCAFFOLD. Importantly, Kuramoto-FedAvg exhibits minimal performance degradation from near-IID to highly non-IID conditions compared to baseline methods, emphasizing its adaptability and robustness to real-world data distributions.

\subsection{Ablation Study: Synchronization Strength \(\kappa_0\)}
\vspace{-0.2cm}

To analyze the role of synchronization strength in Kuramoto-FedAvg, we vary \(\kappa_0 \in \{0.005, 0.1, 0.3\}\) and compare against a no-synchronization baseline across the three benchmark datasets.We report the maximum test accuracy achieved across communication rounds in Table~\ref{tab:full_tuning_results}. This evaluation complements our theoretical claim that synchronization reduces client drift and accelerates convergence, while also revealing dataset-specific sensitivity to the degree of enforced alignment.

Results show that small to moderate synchronization strength improves performance consistently across all datasets. On CIFAR-10, the best performance (73.3\%) is attained with \(\kappa_0 = 0.005\), compared to only 61.42\% without synchronization, indicating that even weak alignment significantly mitigates gradient divergence in high-variance settings. MNIST exhibits more muted gains, likely due to its lower task complexity and gradient diversity; here, \(\kappa_0 = 0.005\) yields a peak accuracy of 98.14\% over 97.67\% in the baseline. On FMNIST, improvements are also evident, with \(\kappa_0 = 0.005\) achieving 85.31\% versus 83.31\% without synchronization. Notably, large values of \(\kappa_0 = 0.3\) degrade performance across all datasets, most severely on FMNIST (dropping to 75.53\%), suggesting that strong synchronization can slow convergence.

These findings demonstrate the importance of careful synchronization tuning: while weak synchronization is often sufficient to improve generalization and stability, excessive coordination (coupling) can be detrimental. Overall, the study highlights \(\kappa_0\) as a flexible control parameter to modulate the balance between global coherence and local diversity in heterogeneous federated settings.

\begin{table}[ht]
\centering
\small
\caption{Maximum test accuracy (\%) and the corresponding round for each \(\kappa_0\) value across datasets.}
\label{tab:full_tuning_results}
\begin{tabular}{llc|c|c|c}
\toprule
\textbf{Method} & \textbf{Param} & \textbf{Value} & \textbf{CIFAR-10/Rounds} & \textbf{MNIST/Rounds} & \textbf{FMNIST/Rounds} \\
\midrule
\multirow{4}{*}{Kuramoto-FedAvg} 
& \( \kappa_0 \)  & 0.005 & \textbf{73.30} / 200 & \textbf{98.14} / 200 & \textbf{85.31} / 200   \\
&                &  0.1  & 72.52 / 200 & 98.10 / 200 & 84.35 / 200  \\
&                &  0.3  & 67.88 / 200 & 98.04 / 200 & 75.53 / 200 \\
&                &  No-sync  & 61.42 / 200 & 97.67 / 200 & 83.31 / 200 \\
\bottomrule
\end{tabular}
\end{table}

\subsection{Discussion}
Empirical results show that Kuramoto-FedAvg effectively mitigates client drift by dynamically weighting client updates based on their alignment with the global gradient, providing a more effective alternative to averaging or correction-based methods such as FedAvg and SCAFFOLD.

Theoretically, Kuramoto-FedAvg reduces the client drift term $\Gamma(t)$ that limits the convergence of FedAvg under heterogeneity . \textbf{Theorem 2} formally proves that tje adaptive weighting of Kuramoto Fed-Avg lowers $\Gamma_{\text{Kuramoto}}(t)$, resulting in fewer communication rounds needed to reach a target accuracy. This advantage is consistently validated across datasets, especially under high heterogeneity (Figures ~\ref{fig:kuramoto_variance} - \ref{fig:kuramoto_accuracy}).

The effectiveness of Kuramoto-FedAvg becomes increasingly pronounced as task complexity and data heterogeneity increase. On MNIST, performance and convergence improvements are modest but consistent, reflecting the stability of updates under simple and skewed distributions. On FMNIST and CIFAR-10, the method yields substantial gains, attributable to its ability to maintain coherent global trajectories despite divergent local objectives. These results underscore the utility of phase-based synchronization in suppressing misaligned updates without requiring modifications to local training or the introduction of auxiliary correction variables. Compared to SCAFFOLD which mitigates drift through variance-reducing control variates at the cost of additional communication and memory overhead, Kuramoto-FedAvg achieves comparable or superior convergence through a lightweight aggregation mechanism.

Kuramoto-FedAvg introduces a theoretically grounded aggregation strategy that leverages synchronization dynamics to directly address gradient misalignment, a core limitation in federated optimization under data heterogeneity. \textbf{Theorem 2} formalizes the reduction in client drift, and empirical results across multiple benchmarks confirm consistent improvements in convergence speed and stability. This shift in focus from local correction to globally coordinated update alignment presents a conceptually distinct approach to mitigating client drift in federated optimization. These results highlight the broader potential of coordination-based methods and motivate further exploration of principled, system-level approaches to managing statistical heterogeneity in federated learning.
\FloatBarrier

\section{Conclusion}
In this paper, we introduced Kuramoto-FedAvg, a synchronization-based federated aggregation method inspired by the mathematical theory of coupled oscillators. Kuramoto-FedAvg successfully mitigates client drift, accelerates convergence, and improves accuracy by dynamically aligning client gradient updates based on their phase alignments with global gradients. Our theoratical analysis establishes rigorous convergence bounds, providing a strong theoretical foundation for the proposed method. Empirical evaluations across benchmark datasets demonstrate clear advantages in variance reduction, convergence acceleration, and accuracy improvements under varying levels of heterogeneity. The performance gains, particularly evident with complex models and highly heterogeneous data, establish Kuramoto-FedAvg as a state-of-the-art federated optimization method. Overall, the synchronization paradigm presented by Kuramoto-FedAvg directly addresses fundamental challenges in FL and opens promising avenues for future research by blending insights from dynamical systems with distributed ML optimization.
\bibliography{main}{}
\bibliographystyle{plain}
\appendix

\section{Proofs}
\subsection{Proof of Theorem 1}
\label{app:proof-theorem1}
\begin{proof}
By the \(L\)-smoothness of \(F(w)\), we have:
\[
F(w_{t+1}) \leq F(w_t) + \nabla F(w_t)^\top (w_{t+1} - w_t) + \frac{L}{2} ||w_{t+1} - w_t||^2.
\]
Using the update rule \( w_{t+1} = w_t - \eta_t \sum_{k=1}^N p_k g_k^t \), this gives:
\[
F(w_{t+1}) \leq F(w_t) - \eta_t \nabla F(w_t)^\top \sum_{k=1}^N p_k g_k^t + \frac{L \eta_t^2}{2} \left|| \sum_{k=1}^N p_k g_k^t \right||^2.
\]
Taking expectation and noting \( \mathbb{E}[g_k^t] = \nabla F_k(w_t) \), we get:
\[
\mathbb{E}[F(w_{t+1})] \leq F(w_t) - \eta_t ||\nabla F(w_t)||^2 + \frac{L \eta_t^2}{2} \mathbb{E} \left|| \sum_{k=1}^N p_k g_k^t \right||^2.
\]
Now decompose the last term:
\[
\mathbb{E} \left|| \sum_{k=1}^N p_k g_k^t \right||^2 = \left|| \sum_{k=1}^N p_k \nabla F_k(w_t) \right||^2 + \mathbb{E} \left|| \sum_{k=1}^N p_k (g_k^t - \nabla F_k(w_t)) \right||^2.
\]
\[
= ||\nabla F(w_t)||^2 + \sum_{k=1}^N p_k^2 \mathbb{E} ||g_k^t - \nabla F_k(w_t)||^2 \leq ||\nabla F(w_t)||^2 + \sigma^2.
\]
Further, since:
\(
\sum_{k=1}^N p_k ||\nabla F_k(w_t)||^2 = ||\nabla F(w_t)||^2 + \Gamma(t),
\)
we conclude:
\[
\mathbb{E}[F(w_{t+1})] \leq F(w_t) - \eta_t ||\nabla F(w_t)||^2 + \frac{L \eta_t^2}{2} \Gamma(t) + \frac{L \eta_t^2}{2} \sigma^2.
\]
\end{proof}

\subsection{Proof of Theorem 2}
\label{app:proof-theorem2}
\begin{proof}
Assuming \(L\)-smoothness and \(\mu\)-strong convexity, FedAvg satisfies the descent inequality:
\[
\mathbb{E}[F(w_{t+1})] \leq F(w_t) - \eta_t ||\nabla F(w_t)||^2 + \frac{L \eta_t^2}{2} \Gamma(t) + \frac{L \eta_t^2}{2} \sigma^2.
\]
where $\Gamma(t) := \sum_{k=1}^N p_k ||\nabla F_k(w_t)||^2 - ||\nabla F(w_t)||^2$
quantifies the heterogeneity-induced client drift.

In Kuramoto-FedAvg, the update is reweighted using phase-based synchronization. The weight assigned to each client is
\[
\rho_k(t)=\frac{\sin\bigl(\bar{\theta}(t)-\theta_k(t)\bigr)}{\sum_{j=1}^{N}\sin\bigl(\bar{\theta}(t)-\theta_j(t)\bigr)},
\]
which favors clients whose gradient directions are close to the global direction (i.e., small \(\bar{\theta}(t)-\theta_k(t)\)). The effective drift then becomes:
\[
\Gamma(t) := \sum_{k=1}^N \rho_k(t)^2 \ ||\nabla F_k(w_t)||^2 - ||\nabla F(w_t)||^2.
\]
Since clients with large divergence are assigned smaller weights, it follows that
\[
\Gamma_{\mathrm{Kuramoto}}(t) < \sum_{k=1}^N p_k ||\nabla F_k(w_t)||^2 - ||\nabla F(w_t)||^2
= \Gamma(t).
\]
Substituting this bound into the descent inequality, we obtain
\[
\mathbb{E}[F(w_{t+1})] \leq F(w_t) - \eta_t ||\nabla F(w_t)||^2 + \frac{L \eta_t^2}{2} \Gamma_{\mathrm{Kuramoto}}(t) + \frac{L \eta_t^2}{2} \sigma^2.
\]
which implies a greater decrease in the expected loss per round as compared to standard FedAvg. Therefore, for any target accuracy threshold \(\epsilon>0\), the number of rounds needed, \(T_{\mathrm{Kuramoto\mbox{-}FedAvg}}(\epsilon,\Gamma)\), satisfies

\[
T_{\mathrm{Kuramoto\mbox{-}FedAvg}}(\epsilon,\Gamma) < T_{\mathrm{FedAvg}}(\epsilon,\Gamma),
\]

which proves that that Kuramoto-FedAvg converges faster under data heterogeneity.
\end{proof}
\FloatBarrier
\section{Additional Figures}
\begin{figure}
  \centering
  \includegraphics[width=1.0\textwidth]{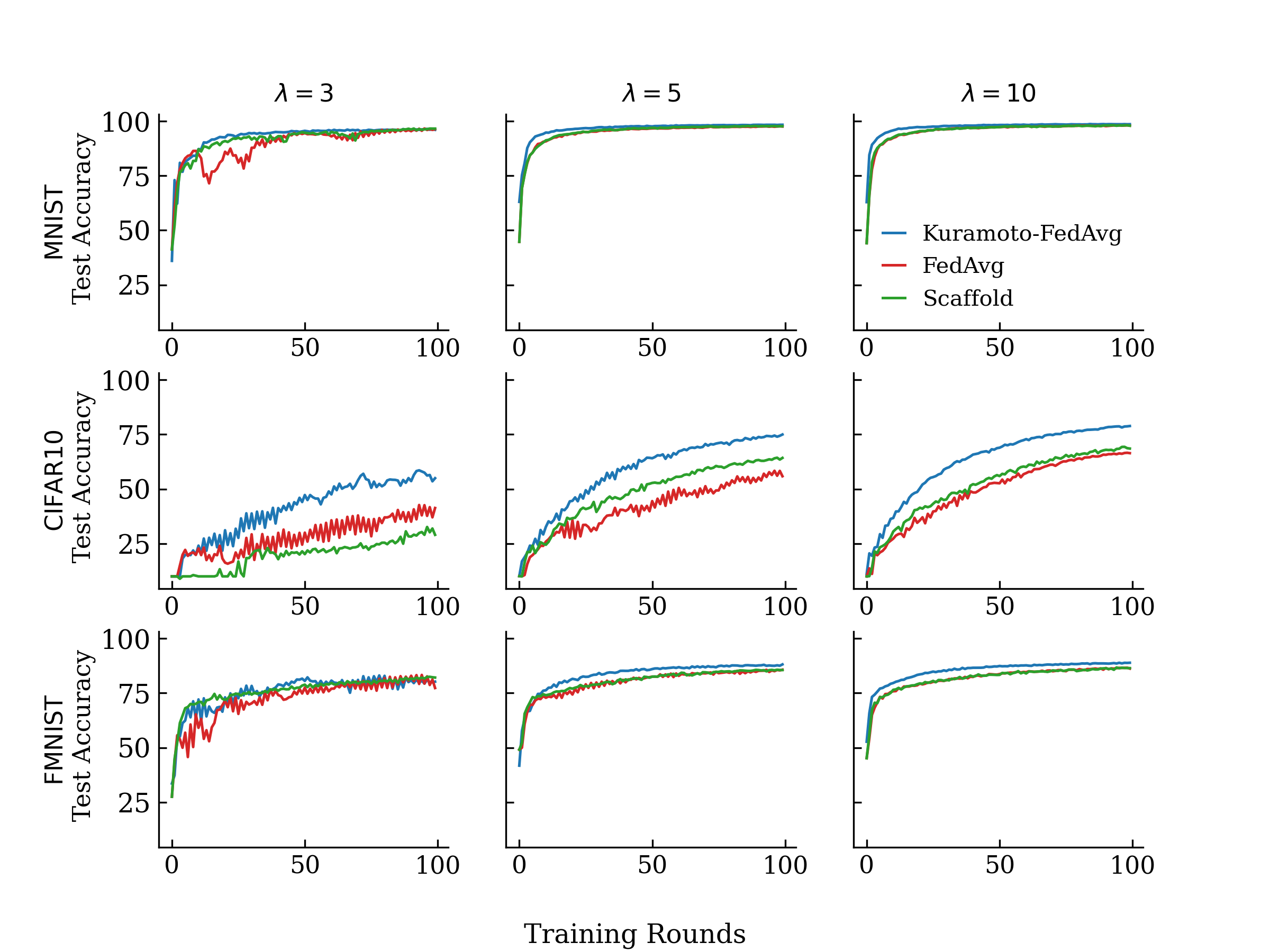}
  \caption{Comparison of test accuracy for Kuramoto-FedAvg, FedAvg, and SCAFFOLD over several training iterations on three benchmark datasets, evaluated at different heterogeneity levels. Kuramoto-FedAvg consistently achieves higher accuracy faster than FedAvg and SCAFFOLD. The rapid accuracy improvement highlights the practical benefits of reducing gradient misalignment through synchronization, demonstrating how Kuramoto-FedAvg effectively addresses model complexity and dataset-specific optimization challenges across varying degrees of client data skewness.}
  \label{fig:kuramoto_accuracy}
\end{figure}
\vspace{0.1em}
\FloatBarrier
\end{document}